\pdfoutput=1

\documentclass{article}
\usepackage{ijcai20}

\usepackage{times}

\usepackage{soul}
\usepackage{url}
\usepackage[hidelinks]{hyperref}
\usepackage[small]{caption}
\usepackage{graphicx}
\usepackage{amsmath}
\usepackage{booktabs}
\usepackage{graphicx}
\usepackage{subcaption}

\usepackage{multirow}
\usepackage{multicol}

\usepackage{amssymb}
\usepackage{amsmath}

\usepackage{mwe}
\title{Hybrid Attention Networks for Flow and Pressure Forecasting in Water Distribution Systems}

\author{
Ziqing Ma $^1$\and
Shuming Liu $^1$\footnote{Contact Author}\and
Guancheng Guo $^1$\and
Xipeng Yu$^{1}$\\
\affiliations
$^1$ School of Environment, Tsinghua University\\
\emails
mazq18@mails.tsinghua.edu.cn,
shumingliu@tsinghua.edu.cn,
\{ggc19, yxp18\} @mails.tsinghua.edu.cn
}

\begin{document}

\maketitle

\begin{abstract}
Multivariate geo-sensory time series prediction is challenging because of the complex spatial and temporal correlation. In urban water distribution systems (WDS), numerous spatial-correlated sensors have been deployed to continuously collect hydraulic data. Forecasts of monitored flow and pressure time series are of vital importance for operational decision making, alerts and anomaly detection. To address this issue, we proposed a hybrid dual-stage spatial-temporal attention-based recurrent neural networks (hDS-RNN). Our model consists of two stages: a spatial attention-based encoder and a temporal attention-based decoder. Specifically, a hybrid spatial attention mechanism that employs inputs along temporal and spatial axes is proposed. Experiments on a real-world dataset are conducted and demonstrate that our model outperformed 9 baseline models in flow and pressure series prediction in WDS.
\end{abstract}

\section{Introduction}
Wide deployment of sensors for systematic monitoring in the physical world is becoming affordable and comprehensive \cite{ekundayo_standardised_2011}. Examples include meteorological sites, traffic monitors and sensors in urban water distribution systems (WDS). These sensors, which monitor one or multiple parameters, continuously generate time series readings. The physical locations of these sensors are fixed in the environment, which results in a certain spatial correlation between these sensors. The readings of these sensors are called geo-sensory time series \cite{liang_geoman:_2018}. The prediction of geo-sensory series is challenging because both spatial and temporal correlations shall be considered.

The scenario discussed in our work is the geo-sensory series in WDS. Short-term flow and pressure forecasting is a hot topic in the research field and efficient models need to be developed. In most cases, operational decision-making, alerts and anomaly detection are based on real-time flow and pressure readings monitored by sensors deployed in WDS \cite{herrera_predictive_2010}. 
A reliable prediction model reveals tendencies in monitored data, which means water supply companies can make informed decisions.
To tackle this problem, two main groups of models were proposed: the mechanism model and the machine learning model. EPANET is a modelling software widely used for simulations of the hydraulic and water quality behavior in water supply systems \cite{m_water_2011_epanet}. However, EPANET does not cope well with unknown boundary conditions. Machine learning models, on the other hand, do not require accurate pipe system boundary conditions and can be trained directly from the historical data. 

\begin{figure}[t]
    \centering
    \begin{subfigure}[b]{0.23\textwidth}
        \centering
        \includegraphics[width=\textwidth]{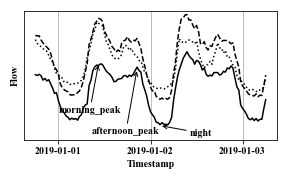}
        \caption[]%
        {{\small Daily pattern of flow}}    
        \label{fig:mean and std of net14}
    \end{subfigure}
    \hfill
    \begin{subfigure}[b]{0.23\textwidth}  
        \centering 
        \includegraphics[width=\textwidth]{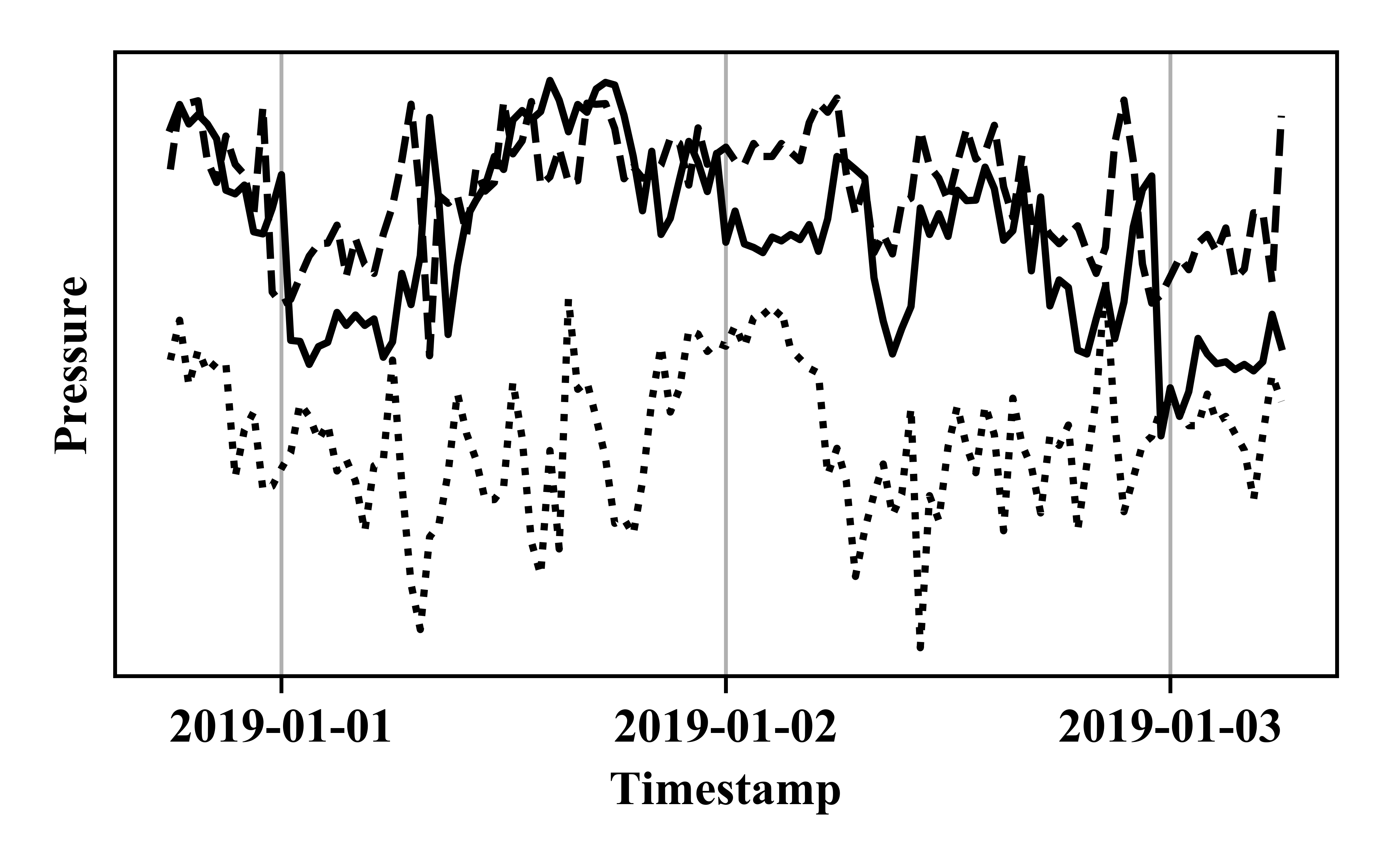}
        \caption[]%
        {{\small Daily pattern of pressure}}    
        \label{fig:mean and std of net24}
    \end{subfigure}
    \caption[]
    {\small The daily pattern of flow and pressure series of 3 different sensors in WDS. The former exhibits daily periodicity while the latter exhibits no obvious regularity.} 
    \label{fig_flow_and_pressure}
\end{figure}

Using machine learning models for flow and pressure prediction in WDS is a challenging work, for the following reasons: (1) Flow and pressure series are nonstationary and difficult to predict. 
Specifically, as shown in Figure \ref{fig_flow_and_pressure}, the flow series exhibits a 24-hour periodic pattern which accounts for the seasonality while the pressure series exhibits no obvious regularity. Moreover, the uncertainty of water consumption behaviour and other unpredictable issue increase the difficulty of the time series forecasting \cite{guo_short-term_nodate}. (2) The pairwise correlation of exogenous series is difficult to address \cite{liang_geoman:_2018}. Normally, in WDS, two hydraulic parameters (pressure and flow) are recorded. The continuity and incompressibility of water ensure the correlation between flow series measured at different junctions of the network. 
Meanwhile, the Bernoulli principle indicates that the variation of the pressure is inverse to the flow. Hence, both exogenous pressure and flow series are of vital importance in the prediction of the target sensor. (3) The capture of temporal correlations is difficult. The pipe network is large so an abrupt event (e.g. Starting a new pump in the booster station) will lead to a response of a remote sensor with an uncertain time lag. 

The complex spatial-temporal correlation of exogenous sensors is difficult to model by traditional machine learning methods  \cite{Qin_dual_stage_attention}. Recent advancement in deep neural networks, especially the employment of attention mechanism, has proven its efficiency in multivariate time series prediction \cite{Qin_dual_stage_attention,liang_geoman:_2018,Yeqi_dual_twophase_attention}. Following previous work, we proposed hybrid spatial-temporal attention-based RNN (hDA-RNN) for the modelling of multivariate time series prediction. The contributions of our work are as follows: 
\begin{enumerate}
    \item To our best knowledge, this work firstly uses recurrent networks with attention mechanism to capture spatial-temporal correlations of flow and pressure series in WDS for their short-term forecasting. 
    \item A hybrid spatial attention mechanism is proposed to enhance the spatial correlation learning in the encoder stage. The proposed method uses inputs along both temporal and spatial axes for the training of attention weighs. Promising results are achieved in the experiments.
    \item Experiments are conducted on a real-world dataset. The proposed model hDA-RNN outperformed 9 baseline models, including the traditional time series model SARIMA and the state-of-the-art attention-based RNN.
\end{enumerate}






\section{Related Work}
Our work focuses on two tasks: the forecasting of flow and pressure series in WDS. 
Short-term water demand forecasting is comprehensively investigated. Because the short-term water demand exhibits very similar pattern to the flow series that we discussed, there are numerous models that we can refer to. 


For short-term water demand forecasting, both traditional models and deep networks are widely investigated. \cite{arandia_tailoring_2016} used seasonal autoregressive integrated moving average model (SARIMA). SARIMA is easy to adapt and requires small training set (7 days) to achieve promising results. Furthermore, using deep learning methods, \cite{guo_short-term_nodate} proposed a deep networks which leverages GRU layers and dense layers and outperformed SARIMA. \cite{msiza_artificial_2007} compared the traditional machine learning model: supported vector regressor (SVR) and the multilayer perceptrons (MLP). A variety of the kernels types of SVR has been tested and promising results were reached. However, the SVR model still underperformed MLp which employs deeper structure and more parameters to extract complex no-linear representation of the inputs\cite{msiza_artificial_2007}. 
The forecasting of pressure series which exhibits no obvious regularity faces more difficulties. Efficient models need to be developed.

Recent advancement of attention mechanism has brought significant improvement in various field. Inspired by the human attention mechanism, the attention-based RNN model was first proposed in a machine translation task \cite{Bahdanau2014NeuralMT}. Employing the encoder-decoder structure and the attention mechanism applied to the encoding sequence, the model became, at that time, the start-of-the-art work in English-to-French translation. Later, the well-known Transformer \cite{attention_is_all_you_need} dispensed with the recurrent module and solely employed several Multi-Head attention modules and feed-forward modules. Proven efficient in sequence to sequence tasks like natural language process \cite{Bahdanau2014NeuralMT,attention_is_all_you_need,yang-etal-2016-hierarchical} and computer vision tasks \cite{Xu_image_caption_with_attention,huang_attention_on_attention_2019}, the attention-based models were also widely used in time series problems. Dual-stage attention-based RNN (DS-RNN) \cite{Qin_dual_stage_attention} employed two attention modules for spatial-temporal correlation capturing and outperformed the other traditional nonlinear autoregressive exogenous models. GeoMAN \cite{liang_geoman:_2018} used three-stage attention respectively applied to local features, global features and temporal sequence for geo-sensory series prediction. 
To enhance the correlation of the target series and exogenous series, a dual-stage two-phase attention model, namely DSTP-RNN \cite{Yeqi_dual_twophase_attention}, was proposed. The first-phase spatial attention was applied to the original series. The second-phase spatial attention was applied to the concatenation of the former output and the target series. The second stage (temporal attention) was the same as the aforementioned works.

However, among the current attention-based RNN models, one essential bottleneck could be the information loss during the training of attention weights. Specifically, the spatial attention mechanism only considers inputs along the temporal axis and ignores exogenous inputs. Hence, we proposed a hybrid spatial attention mechanism in the encoder stage. It could simultaneously capture inputs along temporal and spatial axes to training the attention weights. To the best of our knowledge, no prior work has used a similar method in the spatial attention stage.

\section{Notation \& Problem statement}
Given $n\in \mathbb{N}_+$ series over a window length of $T\in \mathbb{N}_+$ as inputs, we use $x_t^k \in \mathbb{R}$  to represent the reading of k-th sensor at time t. We use $X_t = (x^1_t, x^2_t,..,x^n_t)^T \in \mathbb{R}^{n \times T}$ to represent the encoder input series of all geo-sensors at time t. In addition, the output of the encoder layer is represented by $Z_t = (z^1_t, z^2_t,..,z^m_t)^T \in \mathbb{R}^{m \times T}$, where m is the number of hidden states. The reading of the target sensor at time $t$is represented by $y_t$. Given $\tau$ as the window length of predicted series, the target series for prediction which ranges from $T$ to $T+\tau$ can be represented by $Y=(y_{T+1}, y_{T+2},..., y_{T+\tau})^T \in \mathbb{R}^\tau$. The historical part of target series $y^1_t, y^2_t,..,y^n_t$ is included in input series. We use $\hat{Y}=(\hat{y}_{T+1}, \hat{y}_{T+2},..., \hat{y}_{T+\tau})^T \in \mathbb{R}^\tau$ to represent the predicted series. 

A mapping from historical multivariate series to the future target series is expected to be learned. Note $F( \cdot )$ as the mapping function of the model, the problem is defined as follows:
\begin{equation}
    \hat{y}_{T+1}, \hat{y}_{T+2},..., \hat{y}_{T+\tau} = F(x^1_t, x^2_t,..,x^n_t)
\end{equation}

\section{Methods}
\begin{figure}[t]
    \centering
    \includegraphics[width=0.35\textwidth]{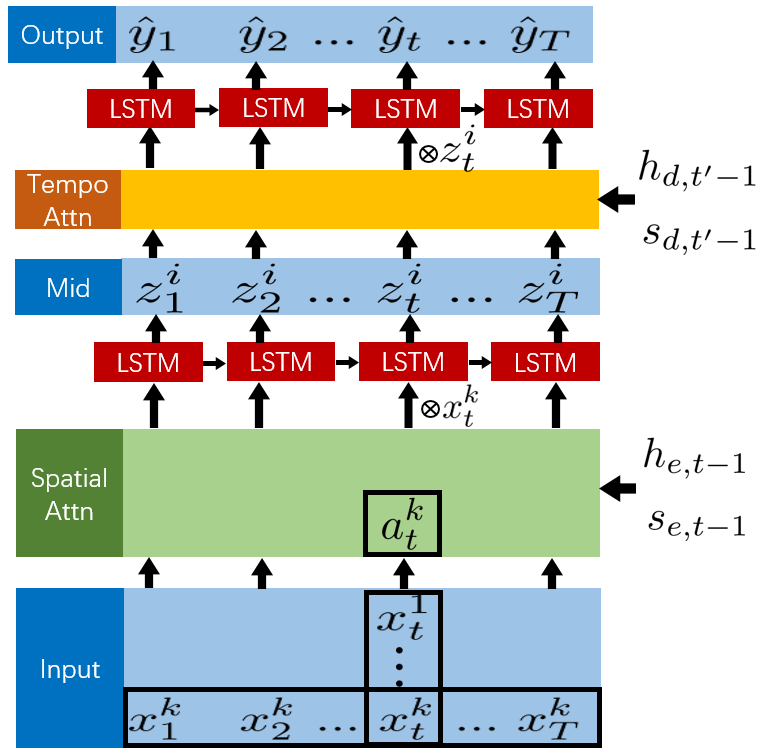}
    \caption{The framework of the proposed hybrid dual-stage attention networks (hDA-RNN). In the first stage, the spatial attention weights $a_t^k$ is trained with the inputs along the temporal axis $(x^k_1, x^k_2,..,x^k_T)^T \in \mathbb{R}^n$ and those along the spatial axis $(x^1_t, x^2_t,..,x^n_T)^T \in \mathbb{R}^n$. In addition, the hidden states and cell states of the previous LSTM unit $h_{e,t-1}$, $s_{e,t-1}$ are employed to train the spatial attention weights. The output of the encoder is represented as $Z_t = (z^1_t, z^2_t,..,z^m_t)^T \in \mathbb{R}^{m \times T}$. In the second stage, the temporal attention computes the attention weights with mid-layer states across all the time steps and the hidden states and the cell states of the decoder $h_{d,t'-1}$, $s_{d,t'-1}$. The output of the decoder is represented as $(\hat{y}_{T+1}, \hat{y}_{T+2},..., \hat{y}_{T+\tau})^T \in \mathbb{R}^\tau$}
    \label{fig_model_structure}
\end{figure}

The overall framework of proposed hybrid DS-RNN is presented in Figure \ref{fig_model_structure}. The encoder-decoder structure \cite{sutskever_sequence_nodate}, proven efficient in sequence to sequence modelling, is employed in our framework. The LSTM unit \cite{hochreiter_long_1997} is used as the basic unit in both encoder and decoder. A dual-stage attention mechanism \cite{Qin_dual_stage_attention} i.e., a spatial attention layer in encoder and a temporal attention layer in decoder, is employed. Specifically, a hybrid spatial attention mechanism is proposed in the first stage. In the second stage, the temporal attention mechanism is employed.

\subsection{Spatial attention in encoder}
The spatial attention mechanism is performed in the first stage. 
Following previous work, 
a dual-stage attention-based RNN (DA-RNN)\cite{Qin_dual_stage_attention} serves as an important baseline in our experiment. Firstly, we present the attention mechanism employed in DA-RNN.  Given $X^k = (x^k_1, x^k_2,..,x^k_3)^T \in \mathbb{R}^T$ as k-th input series and $h_{e,t-1}$, $s_{e,t-1}$ as the hidden states and the cell states of the encoder LSTM cell at step t-1. The spatial attention mechanism can be defined as follows \cite{Qin_dual_stage_attention} \cite{liang_geoman:_2018}
 \cite{Yeqi_dual_twophase_attention}:
\begin{equation}
    e^k_t = V_e \tanh (W_e[h_{e,t-1};s_{e,t-1}] + U_eX^k + B_e)
\label{fun_spa_att}
\end{equation}
\begin{equation}
    a^k_t=\frac{\exp(e^k_t)}{\sum^N_i \exp(e^i_t)}
\end{equation}
where $V_e \in \mathbb{R}^{1 \times l}$, $B_e \in \mathbb{R}^l$ $W_e\in \mathbb{R}^{l \times 2m}$ $U_e\in \mathbb{R}^{l \times T}$ are the parameters to be learned. $l$ is the hidden dimension of the spatial attention module. Specifically, \cite{Qin_dual_stage_attention} used $l=T$. $m$ is the dimension of hidden states: $h_{e,t-1}$ and $s_{e,t-1}$. $[\cdot;\cdot]$ is the concatenation operation. The attention weights $E_t = (e^k_1, e^k_2, ..., e^k_t)^T$ will be treated with a softmax function to make sure the sum equals to 1. 

In the aforementioned work, the spatial attention weights of the k-th sensor $e^k_t$ are trained with the series itself across all time steps $X^k \in \mathbb{R}^T$. The defect of this method is that the pair-wise exogenous correlation is ignored. To overcome this defect, we proposed a modified DS-RNN model, namely DS-RNN-II, which employed a novel spatial mechanism defined as follows:
\begin{equation}
    e^k_t = V_e \tanh (W_e[h_{e,t-1};s_{e,t-1}] + U_eX^t + B_e)
\label{fun_spa_att_II}
\end{equation}
where $V_e \in \mathbb{R}^{1 \times h}$, $B_e \in \mathbb{R}^h$ $W_e\in \mathbb{R}^{h \times 2m}$ $U_e\in \mathbb{R}^{h \times n}$ are the parameters to be learned. $h$ is the hidden dimension of the spatial attention module. Compared with Function \ref{fun_spa_att}, The input series $X^k$ is replaced with $X^t$, which strengthens the influence of the input along spatial axes rather than temporal axes. Two spatial attention mechanism (DS-RNN and DS-RNN-II) will be compared in experiment in the following section.  

Combining the advantages of the two aforementioned spatial attention mechanism, we further develop a hybrid spatial attention mechanism. The model proposed is called hybrid dual-stage attention based recurrent networks (hDA-RNN). As shown in Function \ref{fun_spa_att_III}, the proposed hybrid attention mechanism employed both $X^k$ and $X^t$ to train the spatial attention weights. Considering both the k-th series over a window length of $T$ $X^k$ and the exogenous series at current step t, the attention weight is more suitable. The hybrid attention mechanism is defined as follows:
\begin{equation}
    e^k_t = V_e \tanh (W_e[h_{e,t-1};s_{e,t-1}] + U_eX^k + U'_eX^t + B_e)
\label{fun_spa_att_III}
\end{equation}
\begin{equation}
    a^k_t=\frac{\exp(e^k_t)}{\sum^N_i \exp(e^i_t)}
\end{equation}
where $V_e \in \mathbb{R}^{1 \times h}$, $B_e \in \mathbb{R}^h$ $W_e, L_e\in \mathbb{R}^{h \times 2m}$ $U_e\in \mathbb{R}^{h \times T}$. $h_{e,t-1}$, $s_{e,t-1}$ and $h_{d,t'-1}$, $s_{d,t'-1}$ are the hidden states of the encoder and decoder respectively. In both encoder and decoder layer, we use $m$ as the dimension of hidden states for simplicity.

The output vector of the hybrid spatial attention layer is presented as:
$(a^1_tx^1_t, a^2_tx^2_t, ... , a^n_tx^n_t)^T \in \mathbb{R}^{n}$. The aforementioned output vector is processed by stacked LSTM layer to get the encoder output: $Z_t = (z^1_t, z^2_t,.., z^m_t)^T \in \mathbb{R}^{m \times T}$. The encoder output serves as the input of the temporal attention layer.


\subsection{Temporal attention in decoder}
The performance of the recurrent networks is limited by its 'memory'. The LSTM units enhance the information storage over an extended time interval by using multiplicative gates to regulate the access to its memory \cite{hochreiter_long_1997}. However, the LSTM-based networks still has difficulty in capturing long-term dependency. Hence, by employing a temporal attention mechanism, the capability of capturing long-term dependency was maximised. Because the selectively weighted hidden states along the sequence are connected directly to the decoder, the model suffers no more from the long-term information loss. Taking the output of the spatial attention layer ${Z=(z_1, z_2, ..., z_T) \in \mathbb{R}^{m \times T}}$ and the hidden states of the decoder at time t-1 $h_{d,t-1}$, $s_{d,t-1}$, the dynamic temporal attention weights are defined as follows:
\begin{equation}
    f_{t',t} = V_d \tanh (W_d[h_{d,t'-1};s_{d,t'-1}] + U_dz_t + B_d)
\end{equation}
where $V_d \in \mathbb{R}^{1 \times m}$, $B_d \in \mathbb{R}^m$, $W_d\in \mathbb{R}^{m \times 2}$, $U_d\in \mathbb{R}^{m \times m}$ are the parameters to be trained. The temporal attention weight $f_{t',t}$ represents the importance of the values in encoder at time $t$ for the prediction of the values in decoder at time $t'$. The temporal attention weights are normalised by a softmax function defined as follows:
\begin{equation}
  \beta_{t', t}=\frac{\exp(f_{t', t})}{\sum^T_{j=1} \exp(f_{t', j})}  
\end{equation}
The output of the temporal attention layer is presented as:
\begin{equation}
    \zeta_{t'} = \sum^T_{t=1} \beta_{t', t} Z_t
\end{equation}

\subsection{Encoder Decoder \& model training}
In this section, the encoder-decoder structure and whole training process will be summarised. In the encoder stage, $X^t \in \mathbb{R}^n$ is used as the model input. After processed by spatial attention layer, the middle states are $Z^t \in \mathbb{R}^m$ which will be used as the input of the decoder. Processed by temporal attention layer, the final output of the model $\hat{y}^{t+1}$ is obtained. 

All the deep neural networks use hyperbolic tangent (tanh) function as the activation function to generate no-linear representation defined as follows:
\begin{equation}
    tanh(x)=\frac{\exp(x)-\exp(-x)}{\exp(x)+\exp(-x)}
\end{equation}

The LSTM units are used as the basic elements in the model. LSTM aims at learning a mapping function as follows:
\begin{equation}
    h_t, c_t = f(X_t, h_{t-1}, s_{t-1})    
\end{equation}
where $X_t$ is the input at time t, $h_t$ and $c_t$ are the hidden state and cell state of LSTM unit at time t-1. 

A dropout technique is used to overcome overfitting.

All the deep neural networks are smooth and differentiable, which makes them trainable via backpropagation algorithm   \cite{Rumelhart1986LearningRB}. The mean squared error (MSE) between the predicted series and the ground truth series is minimized by Adam optimizer \cite{kingma_adam:_2017} and the learning rate is selected accordingly.

All deep neural networks are implemented in PyTorch framework. 

\section{Experiments}

In this section, the details of the dataset used will be presented first along with the partition of the training set, validation set and test set. Secondly, a pretreatment process will be presented to transform the unstationary series into stationary series. Then, the evaluation metrics and hyperparameter selection will be detailed. After that, the comparison of the baseline models and the proposed hDS-RNN will be presented along with some remarks. Finally, the experiments for a further explanation of the spatial-temporal mechanism employed are conducted and the results are illustrated.    
\subsection{Datasets}
The experiment was conducted over a real dataset collected by the water supply company of the city of CZ. Sensors deployed in the WDS continuously collected hydraulic parameters (e.g., flow rate and pressure) for systematic monitoring. The dataset ranges from 05/05/2017 to 04/04/2019 with a time interval of 30 minutes. The dataset contains time series of 11 flow rate monitoring sensors and 7 pressure monitoring sensors. The 11 flow sensors and 7 pressure sensors are placed either on the outlet pipes of booster stations or on water mains. The flow sensor No.8 (F8) and pressure sensor No.5 (P5) are respectively regarded as the target sensors. The dataset was partitioned into the training set, validation set and test set. The three sets are split at 09/01/2018 and 01/01/2019, with the scale of approximately 4:1:1.

\subsection{Pretreatment}
The nonstationarity nature of the input series makes prediction difficult. Nonstationary series can be decomposed into $Trend$, $Seasonal$ and $Residual$ \cite{oliveira_parameter_2017}. To simplify the forecasting task, the input series is transformed into its 1st order derivative to remove $Trend$. Then the derivatived series is additively decomposed into $Seasonal$ and $Residual$. 
The decomposition process is performed only on the training set and the seasonal pattern extracted is applied to both validation set and test set. The residual series is the gap between the 1st order derivative series and the seasonal series. Our aim is to predict the residual series to reconstruct the original series.


\subsection{Evaluation metrics \& Hyperparameters}
Mean squared error (MSE) and mean absolute error (MAE) are used to evaluate the models. Their effectiveness has been proved by numerous time-series predicting problems \cite{liang_geoman:_2018}. Note that $y_t \in \mathbb{R}^\tau$ is the ground truth series and $\hat{y}_t \in \mathbb{R}^\tau$ is the predicted series. The two criteria are defined as:
\begin{equation}
    RMSE = \sqrt{\frac{1}{N}\sum^{T+\tau}_{t=T}(\hat{y}_t-y_t)^2}
\end{equation}
\begin{equation}
    MAE = \frac{1}{N}\sum^{T+\tau}_{i=T}(\hat{y}_t-y_t)
\end{equation}
Specifically, because the series is decomposed additively, the MAE and MSE score applied to either residual series or reconstructed series are the same.  

During the training process, the learning rate for the Adam optimiser is 0.001. The mini-batch training is employed for all the deep neural networks. Although small-batch training can improve the generalisation performance, it degrades computational parallelism \cite{masters_revisiting_2018}. 
Considering both the training speed and performance, a series of batch sizes are tested in a preliminary experiment and a batch size of 64 is selected.

The encoder length $T$ is grid searched over \{5, 10, 20, 40, 80\}. Experiments showed that extending encoder length does not improve the performance. As a result, an encoder length of 60 is selected. Considering the error accumulation and interval length of our dataset, the decoder length is empirically set to 4 for future prediction. A decoder length of 4 represents 2 hours in the real world and is long enough for short-term operational decision-making.

Finally, a grid search over \{32, 64, 128, 256\} is conducted for hidden states dimension. Note that the hidden states in the encoder and decoder are the same. Meanwhile, a grid search over \{1, 2, 4\} is conducted for the number of hidden layers. The hDA-RNN model which uses 60 as encoder length, 64 as hidden states dimension and 1 as hidden layer number, outperformed the other parameter settings on the validation set.


\subsection{Model Comparison}
In this section, the performances of the proposed spatial-temporal attention model and the other baseline models are compared. The models are compared on the aforementioned dataset. Flow series (F8) and pressure series (P5) are used as the target series. 
\begin{enumerate}
    \item SARIMA \cite{arandia_tailoring_2016}: Seasonal autoregressive integrated moving average model $(p, d, q) \times (P, D, Q)_s$. We use s=48 because the period is 24 hours and the time interval is 1/2 hour. d = 1 which means 1st order derivative is used to erase the trend component. The other parameters are optimised according to experimental results.
    \item SVR \cite{msiza_artificial_2007}: Supported vector regressor, a traditional machine learning method widely used in regression tasks.
    \item GBRT \cite{friedman_greedy_2000}: An ensemble of simple models (regression tree) proven efficient and easy to adapt in regression tasks.
    \item MLP \cite{koskela_time_nodate_MLP_time_series}: Multilayer perceptrons (MLP), the simplest neural networks for time series problem.
    \item Seq2Seq \cite{sutskever_sequence_nodate}: The sequence to sequence model is originally used in machine translation. Now it is slightly modified for multivariate time series prediction.
    \item DA-RNN \cite{Qin_dual_stage_attention}: DS-RNN uses dual-stage spatial-temporal attention-based RNN in multivariate time series prediction. It was the state-of-the-art model before DSTP-RNN.
    \item DA-RNN-II: A modified dual-stage spatial-temporal attention-based RNN model using the spatial attention mechanism mentioned in Section \ref{fun_spa_att_II}. 
    \item DSTP-RNN \cite{Yeqi_dual_twophase_attention}: DSTP-RNN uses a dual-stage two-phase spatial-temporal attention mechanism and is the state-of-the-art model for multivariate series prediction.
\end{enumerate}

The performance of all models tested is shown in Table \ref{tab_perform}. In terms of flow series prediction, the proposed hybrid spatial-temporal attention model (hDS-RNN) outperforms all baseline models. Remarkably, compared with the state-of-the-art DS-RNN, hDS-RNN model shows 4.1\% and 2.1\% improvements MSE and MAE. DS-RNN-II model uses a modified spatial attention mechanism and improvements on both criteria could be observed. The hybrid DS-RNN model, which is a fusion of DS-RNN and DS-RNN-II, achieves the best performance among them. However, DSTP-RNN model underperforms the DA-RNN model, which doesn't accord with the result in \cite{Yeqi_dual_twophase_attention}. DSTP-RNN-II model in \cite{Yeqi_dual_twophase_attention} is also tested and no obvious improvement is observed. Due to the fact that our dataset is relatively small (contains only 18 series), the large and complex models do not perform well. In the work  \cite{Yeqi_dual_twophase_attention}, results show that the difference of the performance between DA-RNN and DSTP-RNN is small when the dataset is simple. 

In terms of pressure series prediction, compared with the state-of-the-art DA-RNN model, improvements of 1.4\% and 3.3\% on MSE and MAE are achieved. However, Because of the erratic fluctuation of the pressure series, the prediction of pressure faces more difficulties. According to our experimental results, the improvement achieved by hDS-RNN is small compared with traditional time series model SARIMA.





\begin{table}[t]
    \centering
    \begin{tabular}{|c|c|c|c|c|}
    \hline
    \multicolumn{1}{|c}{\multirow{2}{*}{Models}} & \multicolumn{2}{|c|}{FLow} &  \multicolumn{2}{c|}{Pressure} \\
    \cline{2-5}
    & MSE & MAE & MSE & MAE \\
    \hline
    SARIMA      & 3.78 E-3 & 4.18 E-2 & 1.41 E-2 & 8.88 E-2\\
    SVR         & 5.47 E-3 & 4.89 E-2 & 1.77 E-2 & 9.97 E-2\\
    GBRT        & 5.50 E-3 & 4.90 E-2 & 1.72 E-2 & 9.80 E-2\\
    ANN         & 3.44 E-3 & 4.25 E-2 & 1.21 E-2 & 8.32 E-2\\
    Seq2Seq     & 3.28 E-3 & 4.13 E-2 & 1.27 E-2 & 8.48 E-2 \\
    Transformer & 3.09 E-3 & 4.01 E-2 & 1.96 E-2 & 10.4 E-2 \\
    DS-RNN      & 2.17 E-3 & 3.37 E-2 & 1.21 E-2 & 8.30 E-2 \\    
    DS-RNN-II   & 2.12 E-3 & 3.34 E-2 & 1.23 E-2 & 8.46 E-2 \\
    DSTP-RNN    & 2.70 E-3 & 3.72 E-2 & 1.29 E-2 & 8.65 E-2\\
    hDS-RNN     & 2.08 E-3 & 3.30 E-2 & 1.17 E-2 & 8.18 E-2\\ 
    \hline
    \end{tabular} 
    \caption{Performance comparison of different models}
    \label{tab_perform}
\end{table}

\subsection{Evaluation on Temporal attention}
\begin{figure}[t]
    \centering
    \begin{subfigure}[b]{0.23\textwidth}
        \centering
        \includegraphics[width=\textwidth]{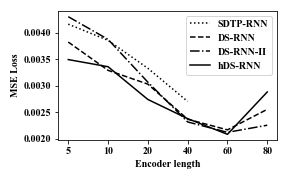}
        \caption[]%
        {{\small Encoder length vs MSE}}    
        \label{fig:mean and std of net14}
    \end{subfigure}
    \hfill
    \begin{subfigure}[b]{0.23\textwidth}  
        \centering 
        \includegraphics[width=\textwidth]{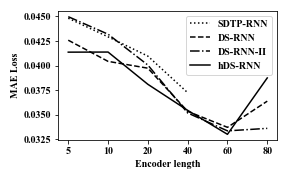}
        \caption[]%
        {{\small Encoder length vs MAE}}    
        \label{fig:mean and std of net24}
    \end{subfigure}
    \caption[]
    {\small Results of encoder window length vs both metrics in water flow prediction of 4 main attention-based RNN (DSTP-RNN, DS-RNN, DS-RNN-II and hDS-RNN).} 
    \label{encoder_len_pred}
\end{figure}

The temporal attention mechanism is employed to capture long-term dependency. In terms of water flow prediction, the Seq2Seq baseline model achieves the best performance when the encoder length is 20. However, the temporal attention based models are capable of capturing longer dependency. Most of them achieve the best performance when encoder length is 60. As presented in Figure \ref{encoder_len_pred}, the performance curves descend steadily as the encoder window extends. After the optimum point T=60, their performance.s drop rapidly. 
Specifically, when $T=100$, DS-RNN model and DS-RNN-II model don't converge in the training process. Moreover, when $T=60$ and $T=80$, DSTP-RNN model fails to converge. The results of no-converged experiments are not shown in Figure \ref{encoder_len_pred}. Suffering from vanishing gradient\cite{pascanu_difficulty_rnn_2013}, the attention-based model still have difficulty in capturing very long-term dependency. Besides, the proposed model hDS-RNN outperformed the stat-of-the-art model DS-RNN and its modified version DS-RNN-II when encoder length is 60.

\subsection{Evaluation on long-term prediction}
\begin{figure}[t]
    \centering
    \begin{subfigure}[b]{0.23\textwidth}
        \centering
        \includegraphics[width=\textwidth]{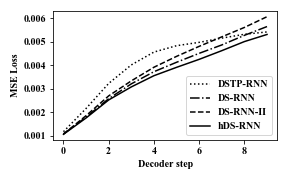}
        \caption[]%
        {{\small Decoder length vs MSE}}    
        \label{fig:mean and std of net14}
    \end{subfigure}
    \hfill
    \begin{subfigure}[b]{0.23\textwidth}  
        \centering 
        \includegraphics[width=\textwidth]{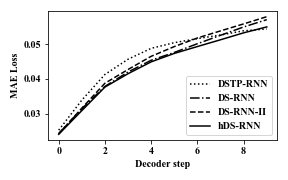}
        \caption[]%
        {{\small Decoder length vs MAE}}    
        \label{fig:mean and std of net24}
    \end{subfigure}
    \caption[]
    {\small Results of decoder window length vs both metrics in water flow prediction of 4 main attention-based RNN (DSTP-RNN, DS-RNN, DS-RNN-II and hDS-RNN).} 
    \label{step_pred}
\end{figure}

Instead of performing the experiment with decoder length $\tau=4$, we extend the decoder length from 4 to 10 for long-term prediction. The capacities of long-term prediction of the four attention-based RNN model (DS-RNN, DS-RNN-II, DSTP-RNN and hDS-RNN) are compared. Both metrics are evaluated on different decoder length of the aforementioned models, as shown in Fig \ref{step_pred}. We observe that the performances of the four models are almost of the same level at step 1 and differ at further steps. Remarkably, the proposed hDS-RNN outperforms the other baseline models at each step. Specifically, the performance of DSTP-RNN trends to surpass its competitors in long-term predictions. In \cite{Yeqi_dual_twophase_attention}, experimental results showed that DSTP-RNN performed better in long-term prediction. However, in our case, we pay more attention on short-term dependency, because in water flow and pressure series prediction, such long dependency is not significant.

\subsection{Evaluation on spatial attention of the model}
\begin{figure}[t]
    \centering
    \begin{subfigure}[b]{0.23\textwidth}
        \centering
        \includegraphics[width=0.826\textwidth]{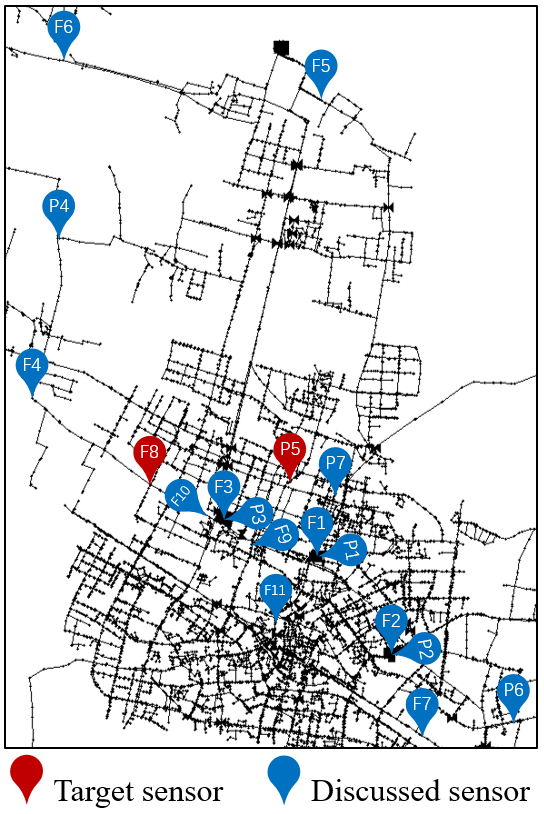}
        \caption[]%
        {{\small Sensor location map.}}    
        \label{fig_map_a}
    \end{subfigure}
    \hfill
    \begin{subfigure}[b]{0.23\textwidth}  
        \centering 
        \includegraphics[width=0.91\textwidth]{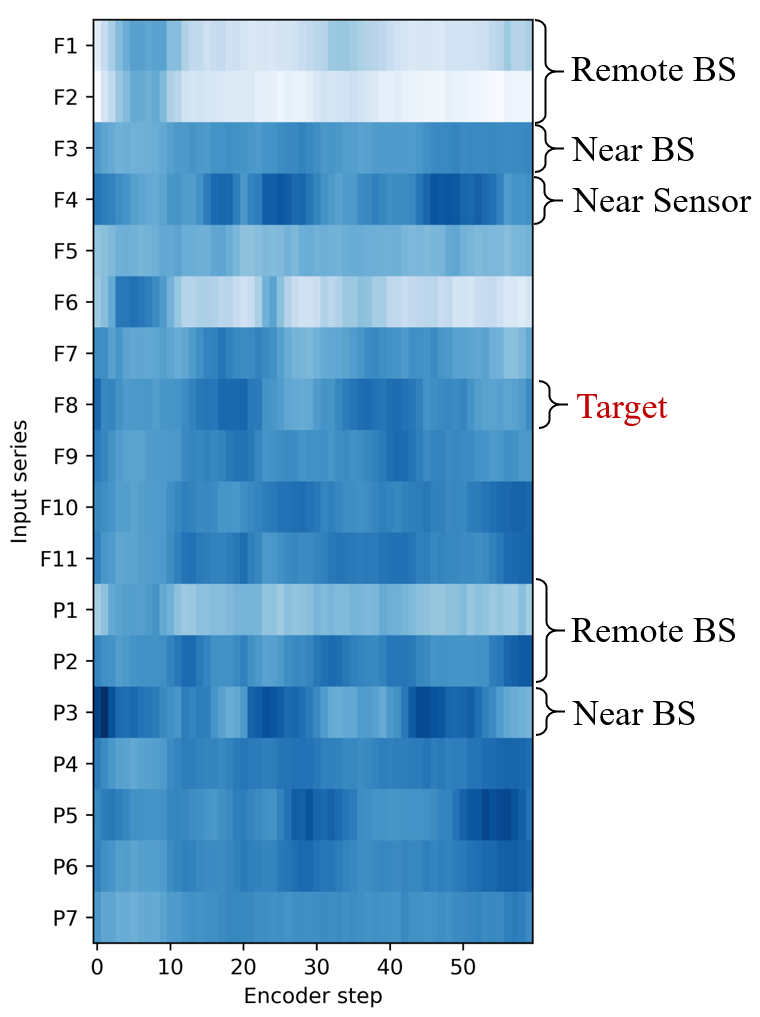}
        \caption[]%
        {{\small Spatial attention weights.}}    
        \label{fig_map_b}
    \end{subfigure}
    \caption[]
    {\small The map of the sensors' location and their attention weights allocated in the model. In this case, 'F8' is the target sensor. 'F1', 'F2', 'P1', 'P2' are recorded on the outlet pipe of two remote booster stations. 'F3', 'P3' are recorded at the nearest booster station. The other series are recorded on water mains} 
\end{figure}

The spatial distribution of geo-sensors are detailed in Figure \ref{fig_map_a}. The spatial attention weights are presented in Figure \ref{fig_map_b}. In this experiment, the target series is F8. It can be observed that the input series F8 is assigned with high weights. The weights assigned to flow series F4 is high, the plausible reason is that their physical locations are close and they are connected directly with a long straight pipe. The weights assigned to series F5 and F6 are relatively small compared with the other flow series because these two sensors are far away from the target sensor F8. The F1 F2 F3 and P1 P2 P3 are the flow and pressure of outflow from three booster station. It can be observed that the spatial attention weights are approximately proportional to the distance. Flow series F3 and pressure series P3, recorded at the nearest booster station, are assigned with higher weights. 

\section{Conclusion and future work}
In this paper, we proposed a hybrid dual-stage attention-based RNN (hDA-RNN) for multivariate time series prediction in WDS. Following previous work, an encoder-decoder structure with dual-stage spatial-temporal attention mechanism was employed. Specifically, we proposed a hybrid spatial attention mechanism in the first stage. In the proposed mechanism, the spatial attention weights of a certain sensor at a certain time is trained with related inputs along temporal and spatial axis simultaneously. Additionally, a temporal attention mechanism is used in the second stage to adaptively allocate different weights for different time steps. Our model is evaluated on a dataset containing hydraulic monitoring series in WDS, respectively using a flow series and a pressure series as the target. Experimental results showed that our model outperformed 9 baseline models on both metrics. Spatial attention weights are visualised for a further explanation of our approach.

In the future, experiments will be conducted on various datasets in different fields . In addition, a real-world application of our model will be conducted.

\section{Acknowledgements}
We would like to thank Dr. Smith for her devoted work in polishing the language. Our work was jointly supported by the National Natural Science
Foundation of China (Grant No. 51879139).



\bibliographystyle{named}
\bibliography{mybib.bib}

\end{document}